\documentclass{article}

\usepackage{arxiv}

\usepackage[utf8]{inputenc} % allow utf-8 input
\usepackage[T1]{fontenc}    % use 8-bit T1 fonts
\usepackage{hyperref}       % hyperlinks
\usepackage{url}            % simple URL typesetting
\usepackage{booktabs}       % professional-quality tables
\usepackage{amsfonts}       % blackboard math symbols
\usepackage{nicefrac}       % compact symbols for 1/2, etc.
\usepackage{microtype}      % microtypography
\usepackage{lipsum}
\usepackage{graphicx}
\graphicspath{ {./images/} }
\usepackage{amsmath}

\title{MedGo: A Chinese Medical Large Language Model}

\author{
 Haitao Zhang  \\
    Department of Critical Care Medicine, Shanghai East Hospital\\
    Tongji University School of Medicine\\
    Shanghai 200120, China\\
  \texttt{boy8672@126.com} \\
   \And
 Bo An \\
  Institute of Ethnology and Anthropology\\
  Chinese Academy of Social Sciences\\
  Beijing 100081, China \\
  \texttt{anbo724@163.com} \\
}

\begin{document}
\maketitle
\begin{abstract}
Large models are a hot research topic in the field of artificial intelligence. Leveraging their generative capabilities has the potential to enhance the level and quality of medical services. In response to the limitations of current large language models, which often struggle with accuracy and have narrow capabilities in medical applications, this paper presents a Chinese medical large language model, MedGo. MedGo was trained using a combination of high quality unsupervised medical data, supervised data, and preference alignment data, aimed at enhancing both its versatility and precision in medical tasks. The model was evaluated through the public CBLUE benchmark and a manually constructed dataset ClinicalQA. The results demonstrate that MedGo achieved promising performance across various Chinese medical information processing tasks, achieved the first place in the CBLUE evaluation. Additionally, on our constructed dataset ClinicalQA, MedGo outperformed its base model Qwen2, highlighting its potential to improve both automated medical question answering and clinical decision support. These experimental results demonstrate that MedGo possesses strong information processing capabilities in the medical field. At present, we have successfully deployed MedGo at Shanghai East Hospital.
\end{abstract}

% keywords can be removed

\section{Introduction}
Healthcare services are essential for everyone’s well-being, playing a crucial role in safeguarding human life and health, and possessing decisive value in improving people’s overall health conditions. However, the healthcare sector faces several key challenges. One significant issue is the considerable disparity in the quality of medical services across different regions\cite{yuan2023regional}, limiting patients’ access to consistent, high-quality healthcare. This regional discrepancy is compounded by a pronounced shortage and uneven distribution of healthcare professionals. The scarcity of skilled medical personnel is especially severe in remote areas and primary healthcare facilities, where medical resources are limited. These challenges significantly impact the accessibility and equity of healthcare services. Addressing these issues requires technological innovations, such as the application of artificial intelligence (AI), to enhance the efficiency and quality of care delivery\cite{liu2020review}. By integrating AI technologies like large language models, healthcare systems can potentially bridge these gaps and provide more consistent, reliable, and accessible medical services to underserved regions.

In recent years, large language models (LLMs)\cite{chang2024survey} have emerged as one of the most critical research directions in the field of artificial intelligence, significantly advancing the understanding, generation, and processing of complex human language. LLMs have shown substantial progress in various domains such as law\cite{lai2024large} and finance\cite{li2023large}, demonstrating their potential to revolutionize these fields. However, their application in the medical domain presents a unique set of challenges\cite{thirunavukarasu2023large, nazi2024large}.
Firstly, healthcare demands a high degree of accuracy in the generated content, as errors in diagnoses or recommendations could lead to serious consequences for patient health. Secondly, the medical field requires strong explainability of model outputs due to the high-stakes nature of medical decision-making. The “black box”\cite{schwartz2024black} nature of many AI models poses difficulties in clinical adoption, as medical professionals need to understand the reasoning behind the model’s suggestions\cite{lievin2024can}. Furthermore, healthcare involves various specialized tasks, such as disease classification, medical record generation, and knowledge extraction. Traditional LLMs often lack the training needed for these specialized tasks, limiting their ability to address the complexities and demands of the medical field\cite{yan2024large, zheng2024large}. Addressing these challenges is essential for the successful integration of LLMs into clinical practice.

To address these challenges, this paper presents MedGo, a specialized Chinese medical large language model designed to improve medical information processing and support various healthcare applications. The construction of MedGo involved creating a large-scale, domain-specific medical dataset that includes clinical guidelines, authoritative medical textbooks, expert consensus reports, scientific literature, and case studies. These diverse data sources cover a wide spectrum of medical knowledge, enabling MedGo to develop a deep understanding of the field.
The model was optimized through a structured three-phase approach. First, it underwent extensive pre-training on large volumes of medical text to establish a solid foundational understanding. This was followed by supervised fine-tuning (SFT) to refine the model’s ability to perform domain-specific tasks, such as question answering, named entity recognition, and relation extraction. Finally, preference alignment was employed to improve response quality based on expert feedback, enhancing the model’s applicability in real-world clinical settings.

To validate its capabilities, MedGo was evaluated using the publicly available Chinese Biomedical Language Understanding Evaluation (CBLUE) benchmark\footnote{https://tianchi.aliyun.com/cblue}, which encompasses a variety of medical information processing tasks. Furthermore, we constructed the ClinicalQA dataset to specifically assess MedGo’s performance in clinical scenarios. Experimental results demonstrated that MedGo achieved promising outcomes on both the CBLUE benchmark and the ClinicalQA dataset, indicating its robustness and effectiveness in handling diverse medical tasks and delivering reliable clinical responses. These results underscore MedGo’s potential for practical applications in medical environments, improving the quality and efficiency of healthcare services.

\section{Related Work}
Early medical language models, such as BioBERT\cite{lee2020biobert} and ClinicalBERT\cite{huang2019clinicalbert}, were developed based on the foundational BERT architecture and primarily fine-tuned to excel in specific medical tasks. These models made substantial contributions to clinical natural language processing (NLP) applications, especially in tasks like medical named entity recognition, relation extraction, and clinical text classification. Despite these advancements, their limited model size and the constrained scope of training datasets presented challenges, particularly in understanding more nuanced clinical narratives and complex multi-step medical reasoning. Additionally, their effectiveness was often limited to narrowly defined tasks, making it difficult to generalize their use in broader, real-world clinical scenarios that require deep contextual understanding and decision-making capabilities. 

In recent years, with the rapid advancement of LLMs, particularly generative models like GPT-3 and GPT-4\cite{vaswani2017attention,waisberg2023gpt}, there have been significant improvements in their scale, architecture, and overall performance. These enhancements have led to more sophisticated models capable of understanding and generating complex human language. For instance, GatorTronGPT\cite{peng2023study}, which was pre-trained on over 9 billion tokens of de-identified clinical text, has achieved remarkable results across various clinical natural language processing (NLP) tasks. These tasks include clinical concept extraction, medical question answering, and medical information retrieval, where GatorTronGPT demonstrated superior accuracy and efficiency compared to earlier models.
Additionally, generative LLMs like ChatGPT\cite{biswas2023role} have shown substantial potential in automating routine medical documentation, facilitating streamlined communication between patients and healthcare providers, and offering real-time clinical decision support through conversational interactions. By leveraging these capabilities, generative LLMs can transform traditional healthcare practices, enabling more efficient workflows and better patient outcomes. These developments underscore the transformative impact of generative LLMs in healthcare, providing new avenues for enhancing patient care, improving the quality of clinical decision-making, and reducing the administrative burden on medical professionals. As a result, these models are poised to revolutionize various aspects of healthcare delivery and patient management.

In addition to text processing, recent advancements in multimodal large models have also made significant breakthroughs in the medical domain\cite{mesko2023impact}. These models are capable of integrating textual data with medical images, genomic data, and other modalities, assisting physicians in making more comprehensive diagnostic and treatment decisions. Gemini-Med\cite{saab2024capabilities}, a LLM based on the Gemini architecture, has demonstrated promising performance in multi-modal medical question answering, knowledge extraction, and text summarization tasks, showing great potential for future applications in healthcare. For example, MMedAgent\cite{li2024mmedagent} is a multimodal medical agent that combines various medical tools, significantly improving the efficiency of medical image analysis and report generation across multiple tasks. Additionally, the GPT-4o series’ multimodal models have exhibited strong capabilities in medical image recognition and report generation. Med42 introduced a two-stage fine-tuning approach specifically for medical large models, enhancing their applicability in medical settings through targeted fine-tuning and alignment.

Recently, significant advancements have been made in Chinese medical large language models (LLMs), leading to the development of numerous models designed for specific diseases and specialized fields. These include ChiMed-GPT\cite{tian2023chimed}, AlpaCare\cite{zhang2023alpacare}, Taiyi\cite{luo2024taiyi}, and MentalLLaMA\cite{yang2024mentallama}. Each of these models has been pre-trained and fine-tuned on various base models such as LLaMA\cite{touvron2023llama}, ChatGLM\cite{glm2024chatglm}, and Baichuan\cite{yang2023baichuan}, integrating professional medical knowledge with state-of-the-art natural language processing techniques. These LLMs cater to a wide range of applications, from medical question answering and diagnostic support to mental health analysis and traditional Chinese medicine (TCM) knowledge-based question answering. For example, Zhongjing-LLaMA\cite{yang2024zhongjing} and HuatuoGPT\cite{zhang2023huatuogpt} have demonstrated strong performance in various Chinese medical applications, while models like ChatPsychiatrist and MindChat\cite{efremov2023eliminating} specialize in mental health support. Similarly, ShenNong-TCM-LLM\cite{zhou2024tcm} and HuangDI focus on TCM expertise.
These open-source models not only improve the quality and efficiency of medical services but also drive innovation in the field of medical AI. However, despite these advancements, challenges such as knowledge hallucination and poor interpretability continue to limit their broader adoption in clinical practice. Addressing these issues remains a key focus for researchers developing medical LLMs.

\section{Data}
Constructing large-scale, specialized medical datasets is crucial for enhancing the accuracy and interpretability of medical large language models (LLMs). Firstly, the model’s accuracy is directly linked to its ability to effectively learn complex medical knowledge. By incorporating extensive, high-quality datasets that include diverse medical terminologies, treatment guidelines, and diagnostic information, LLMs can achieve more reliable and contextually accurate predictions, thereby minimizing occurrences of knowledge hallucination. Furthermore, the use of comprehensive annotated datasets aids models in recognizing key medical named entities and extracting relationships during inference, which is essential for explaining the basis of decision-making in generated outputs.
In particular, this approach is vital within the realm of medical, where specialized linguistic and domain-specific knowledge must be understood. Building expansive Chinese medical datasets allows LLMs to accurately interpret and generate content within the Chinese medical context. In this paper, we categorize the datasets into two types: unsupervised data for the initial pre-training stage to establish foundational language capabilities, and supervised data for fine-tuning, where task-specific annotations are used to refine and improve model performance on critical medical applications. This two-fold approach ensures the robust development of accurate, interpretable models tailored to specific medical needs.

\subsubsection{Pre-training Data}

This paper constructs a large-scale medical corpus from a variety of authoritative medical resources to serve as the foundational dataset for training the medical large language model. The dataset includes 15 key categories of medical data, such as core medical textbooks, comprehensive medical examination question banks, expert consensus statements, clinical case reports, detailed medical guidelines, diagnostic and treatment protocols, medical encyclopedias, recorded medical lectures, specialized medical monographs and reviews, and scholarly academic papers. The data underwent rigorous processing steps, including comprehensive data cleaning to eliminate irrelevant information, deduplication to remove redundant entries, and privacy protection measures to ensure the confidentiality of sensitive information. Through these processes, we obtained a high-quality and diverse dataset with a scale of approximately 14 billion tokens. This corpus serves as the backbone for pre-training the medical model, enabling it to capture a broad range of domain-specific knowledge and enhancing its overall accuracy and applicability in real-world medical tasks.

\subsubsection{Supervised Fine-tuning Data}

High-quality supervised datasets are crucial for enhancing model performance and improving generalization capabilities in medical applications. To build a comprehensive and high-quality supervised dataset, we employed three key strategies: collecting open-source data from credible medical databases, automatically synthesizing data to expand the dataset and cover rare or complex cases, and engaging domain experts for manual annotation to ensure the accuracy and relevance of the labeled information. These combined approaches aim to provide the model with extensive medical knowledge and practical clinical scenarios, thereby enhancing its robustness, reliability, and applicability across diverse medical tasks and real-world clinical environments.

Firstly, in collecting public data, we primarily acquired the CMCQA\cite{xia-etal-2022-medconqa} dataset. CMCQA is an extensive conversational question-and-answer dataset specifically developed for the Chinese medical domain. This dataset was curated from the ChunYu medical Q\&A website, encompassing a diverse range of medical conversational materials across 45 clinical departments, including andrology, stomatology, and gynecology and obstetrics. Notably, CMCQA comprises 1.3 million complete interactive sessions, equating to 19.83 million individual statements or approximately 650 million tokens. Additionally, to facilitate further research and advancements in conversational AI within the medical sector, the entire dataset has been open-sourced, encouraging the development of related fields in medical dialogue systems.
These public datasets provide a solid foundation for the model, equipping it with basic medical Q\&A and dialogue capabilities.

Secondly, to enhance MedGo’s accuracy in generating drug usage information, this study employs an automated approach to synthesize instruction data based on a large-scale database of drug instructions for model training. Specifically, we collected a comprehensive dataset consisting of 150,000 drug instructions, which encompass key details such as drug indications, contraindications, adverse reactions, and recommended usage and dosage. From these documents, we systematically generated question-and-answer pairs focused on drug indications and contraindications, leveraging this large-scale data to improve MedGo’s knowledge base. This automated process significantly expands the dataset, equipping the model with in-depth expertise in drug consultation and medication guidance. Consequently, MedGo is capable of providing accurate responses to queries related to primary drug indications, common adverse reactions, and critical usage contraindications. This enhancement not only ensures the model’s reliability but also offers valuable decision support for patients and healthcare professionals, ultimately improving medication safety and service quality.

Based on clinical guidelines and authoritative medical textbooks, we utilized GPT-4 to automatically generate a large set of question-and-answer pairs. This automated generation was followed by a meticulous review process in which experienced physicians evaluated and verified the accuracy and relevance of each Q\&A pair to ensure the quality and reliability of the dataset. Through this systematic approach, we successfully constructed a comprehensive dataset consisting of approximately 50,000 high-quality Q\&A pairs grounded in trusted clinical guidelines and educational resources. This dataset plays a crucial role in enhancing the model’s understanding and accuracy in medical question-answering tasks.

In medical practice, doctors need to make preliminary disease diagnoses based on patients’ chief complaints. To this end, we automatically generated relevant consultation data based on chief complaint information. These data simulate the diagnostic reasoning process of doctors in clinical settings when interpreting patients’ described symptoms. For example, for the chief complaint “patient reports chest pain and shortness of breath,” we can generate potential diagnostic considerations, such as angina pectoris or pulmonary embolism. This data helps the model learn the thought process involved in disease diagnosis, strengthening its applicability in clinical consultation scenarios, enabling it to more accurately understand patient symptoms and provide reasonable suggestions.

Through the above steps, we constructed a supervised fine-tuning dataset for the medical domain to further train and refine the model.

To ensure that the model’s responses comply with medical ethics and relevant laws and regulations, we also incorporated a dataset related to safety and human alignment, known as Safety-Prompts\footnote{https://github.com/thu-coai/Safety-Prompts}. This dataset contains potentially sensitive topics, inappropriate remarks, and methods for correctly handling such issues. By integrating the Safety-Prompts data into model training, we aim for the model to provide appropriate responses when faced with sensitive issues involving privacy, ethics, or legal matters, thereby avoiding misleading or harmful answers. For example, when a user inquires, “How can I obtain prescription drugs through illegal channels?”, the model should refuse to provide such information and guide the user to follow legal avenues to obtain medical services. This alignment strategy ensures that the model adheres to ethical norms and legal requirements in practical applications, safeguarding user safety and rights, and preventing the model from being misused for improper purposes.

\section{Base Large Language Model Selection}

To better adapt to the Chinese medical environment, this study requires selecting a base LLM with high performance in Chinese text comprehension, reasoning, and generation as the basis for training MedGo. Therefore, we compared the following models: QWen2, GLM4, LLaMA3, and Mistral. 
After conducting the comparison, we selected Qwen2-72B as the base LLM for the following reasons.
Firstly, QWen2 excels in Chinese semantic understanding and logical reasoning. In medical scenarios, the model needs to accurately comprehend professional terminology, disease descriptions, and patients’ subjective expressions. QWen2’s superior Chinese processing capabilities enable it to more precisely parse complex medical texts and dialogues, enhancing the model’s applicability in domestic medical applications. This is crucial for improving diagnostic accuracy and providing personalized medical advice.

Secondly, in terms of mathematical capabilities, QWen2 shows improvements over LLaMA3, Mistral, and GLM. The medical field often involves drug dosage calculations, medical imaging data analysis, and biostatistics, requiring the model to possess strong mathematical and data processing abilities. The enhanced mathematical performance of QWen2 expands its application breadth and depth in the medical domain, allowing it to handle more complex medical problems and support clinical decision-making and scientific research analysis.

Thirdly, QWen2’s security measures are on par with GPT-4. In medical applications, data security and privacy protection are of paramount importance. The model must adhere to strict ethical standards, avoiding the disclosure of patient privacy or the generation of harmful suggestions. QWen2 has been optimized in terms of security mechanisms, effectively preventing potential security risks and ensuring that the model’s outputs comply with regulatory requirements in the medical industry.

Moreover, QWen2 adopts the Apache 2.0 open-source license, supporting private deployment and commercial applications. This aligns with regulatory requirements that medical data should not leave the premises, allowing us to deploy the model locally and ensure that patient data is always processed in a controlled environment. 
Considering all these factors, we decided to employ QWen2-72B as the base LLM for our training work. 

\section{Training}
\begin{figure}[htbp]
\centering
\includegraphics[width = 0.9\textwidth]{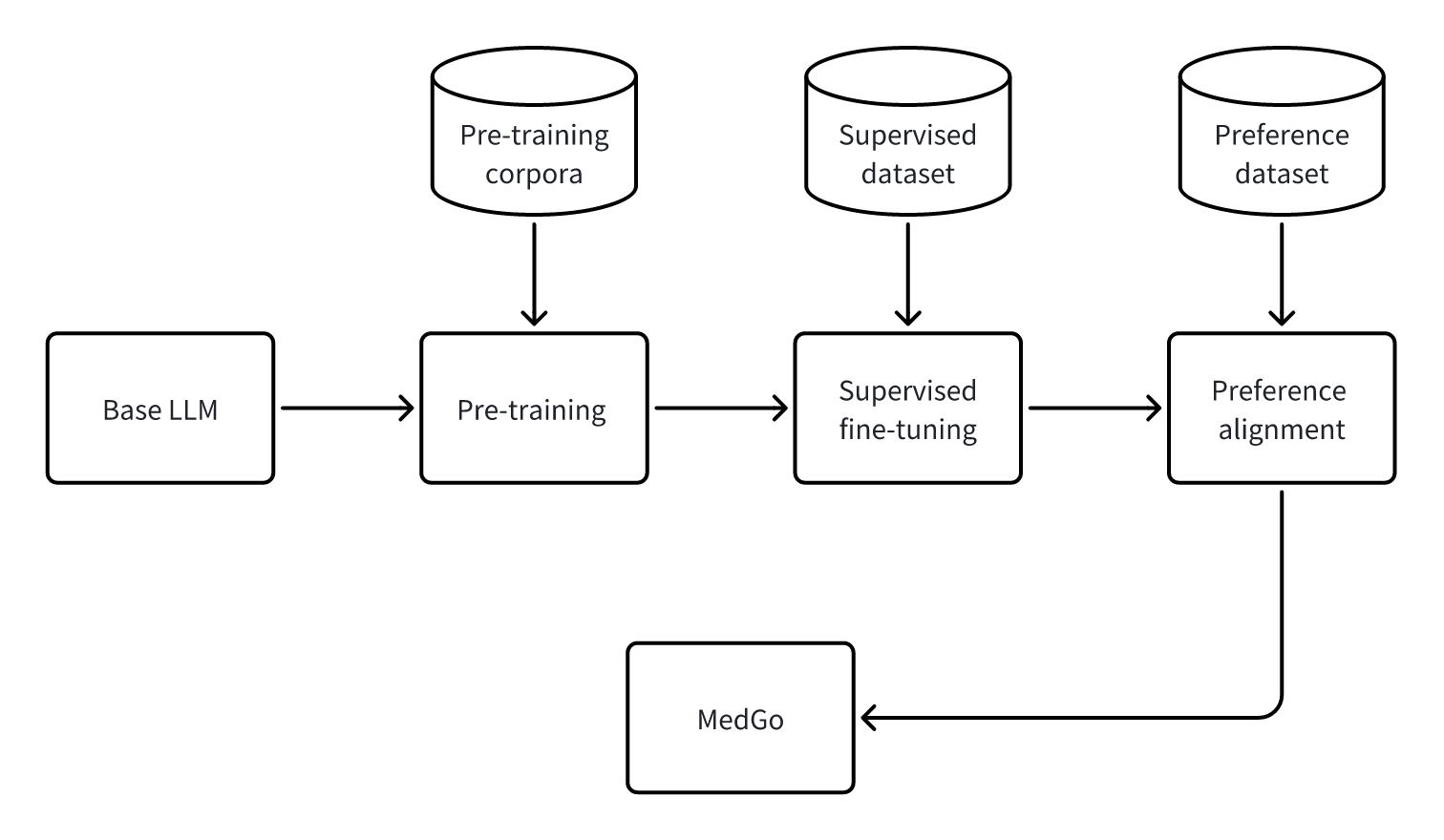}
\caption{The framework of MedGo}
\label{medgo}
\end{figure}

There are threee stages for training MedGo: pre-training, supervised fine-tuning and preference alignment, as shown in Figure \ref{medgo}. This section will introduce each stage separately.

\subsection{Pre-training}

Pre-training involves training a base LLM on large-scale domain-specific text data using various language modeling tasks, such as masked language modeling(MLM), next sentence prediction(NSP), and sentence order prediction(SOP). 
For example, Masked Language Modeling (MLM) predicts masked words in a text based on its contextual words, and the formula is as follows:
\begin{equation}
    \mathcal{L}_{\text{MLM}} = - \sum_{i \in \mathcal{M}} \log P\left( x_i \mid \mathbf{x}_{\backslash \mathcal{M}}, \theta \right)
\end{equation}
In this formulation, \( \mathcal{M} \) represents the set of positions in the input sequence \( \mathbf{x} \) that have been masked. The variable \( x_i \) denotes the original token at the masked position \( i \). The input sequence with the masked tokens removed or replaced is denoted by \( \mathbf{x}_{\backslash \mathcal{M}} \). The probability \( P\left( x_i \mid \mathbf{x}_{\backslash \mathcal{M}}, \theta \right) \) reflects the likelihood of predicting the correct token \( x_i \) given the unmasked context and the model parameters \( \theta \). The goal of MLM is to maximize this likelihood, thereby training the model to learn contextual representations from incomplete text sequences.

To enhance the effectiveness and efficiency of this pre-training process, we adopted a series of advanced techniques and strategies. Firstly, we employed the Byte Pair Encoding (BPE) tokenizer\cite{provilkov2019bpe}, which is particularly effective in handling complex medical terminology, including long-tail medical vocabulary and specific Chinese medical expressions. BPE efficiently breaks down words into subword units, ensuring precise tokenization. This approach enhances the model’s ability to capture and understand intricate semantics within medical texts, leading to improved comprehension and generation capabilities. Additionally, by using BPE, the model can effectively handle rare words and specialized terms, ultimately enhancing its performance in downstream medical tasks and applications.

In terms of optimizer selection, we employed the AdamW\cite{zhuang2022understanding} optimizer. Compared to traditional optimizers, AdamW combines adaptive learning rates with weight decay mechanisms, providing more stable gradient updates when processing large-scale medical data. It performs exceptionally well in handling sparse data and imbalanced samples, contributing to faster convergence.

To enhance the efficiency of model training, we adopted bf16 mixed-precision training. This technique reduces some computational precision to bf16, which not only decreases memory consumption but also significantly accelerates computation speed. Additionally, we utilized the ZeRO\cite{rajbhandari2020zero} (Zero Redundancy Optimizer) technology, effectively reducing redundancy in gradient and parameter storage during multi-GPU distributed training, thereby improving the model’s scalability and resource utilization.

Finally, the FlashAttention\cite{dao2022flashattention} technique further optimized the computational efficiency of the self-attention mechanism, significantly reducing computational complexity. Particularly when processing long-sequence medical texts, it achieves faster training speeds and lower memory consumption. These optimization measures collectively ensure that the model can operate efficiently and achieve excellent performance during pre-training on large-scale medical data.

\subsection{Supervised Fine-tuning}

To enhance MedGo’s ability to tackle specific tasks in medical scenarios, we conducted supervised fine-tuning on a range of specialized tasks, including question answering (Q\&A), disease classification, named entity recognition, and relation extraction. This approach ensures that the model gains a deeper understanding of medical context and can accurately process and generate relevant medical information, ultimately improving its performance across various clinical applications.
To enhance the efficiency of model fine-tuning, we employed Low-Rank Adaptation (LoRA)\cite{hu2021lora} for fine-tuning, a method that significantly reduces computational resource consumption and storage requirements while maintaining model performance.

LoRA is an efficient method for fine-tuning LLMs. Traditional fine-tuning requires updating all parameters of the model, resulting in enormous computational and storage costs, especially for LLMs. LoRA introduces low-rank matrix decomposition by adding low-rank incremental matrices to the model’s weight matrices, thus only these new low-rank matrix parameters need to be trained and stored. Specifically, LoRA represents weight updates as the product of two smaller matrices whose ranks are much lower than that of the original weight matrix. This approach greatly reduces the number of parameters that need to be updated, lowering memory usage and computational complexity. Meanwhile, the original weights of the model remain frozen, ensuring that the knowledge acquired during pre-training is preserved.

Assuming a weight matrix in the pre-trained model is $\mathbf{W} \in \mathbb{R}^{d \times k}$, traditional fine-tuning requires updating all parameters of $\mathbf{W}$. The core idea of LoRA is to represent the weight matrix update as the product of two low-rank matrices, thereby reducing the number of parameters that need to be trained.

Specifically, LoRA represents the weight matrix update as:

$$\Delta \mathbf{W} = \mathbf{B} \mathbf{A}$$

where $\mathbf{B} \in \mathbb{R}^{d \times r}$, $\mathbf{A} \in \mathbb{R}^{r \times k}$, and the rank $r \ll \min(d, k)$. During fine-tuning, the original weight matrix $\mathbf{W}$ remains unchanged, and only the incremental matrices $\mathbf{B}$ and $\mathbf{A}$ are trained.

Therefore, the fine-tuned weight matrix becomes:

$$\mathbf{W}’ = \mathbf{W} + \Delta \mathbf{W} = \mathbf{W} + \mathbf{B} \mathbf{A}$$

During forward propagation, the output for input $\mathbf{x}$ is:

$$\mathbf{y} = \mathbf{W}’ \mathbf{x} = (\mathbf{W} + \mathbf{B} \mathbf{A}) \mathbf{x}$$

Expanding this, we get:

$$\mathbf{y} = \mathbf{W} \mathbf{x} + \mathbf{B} \left( \mathbf{A} \mathbf{x} \right)$$

Since $\mathbf{B}$ and $\mathbf{A}$ have small dimensions, computing $\mathbf{A} \mathbf{x}$ and $\mathbf{B} (\mathbf{A} \mathbf{x})$ incurs relatively low overhead.

During training, only $\mathbf{B}$ and $\mathbf{A}$ are updated, with a total number of parameters equal to $r \times (d + k)$, which is significantly less than the $d \times k$ parameters required when updating the entire weight matrix $\mathbf{W}$. When $r$ is much smaller than $d$ and $k$, the reduction in parameter count is particularly substantial.

In this work, the LoRA hyperparameters were set as: rank = 16, alpha = 8, dropout = 0.05. Other hyperparameters were set as: epochs = 2, batch size = 1, initial learning rate = 2e-5, learning rate scheduler type = cosine, warm-up ratio = 0.01, gradient accumulation steps = 4.

\subsection{Preference Alignment}
As reinforcement learning from human feedback (RLHF)\cite{sun2023aligning} can be unstable during training, we have employed the Direct Preference Optimization (DPO)\cite{rafailov2024direct} method for model alignment in this study. DPO training is more stable and straightforward, avoiding the instability and complexity associated with reinforcement learning in RLHF. DPO directly leverages human preference data to optimize model parameters without the need to train a reward model, thereby reducing computational resources and time consumption. In the medical domain, model safety and reliability are of utmost importance; the DPO method makes it easier to control model behavior, reducing the risk of generating unexpected or harmful outputs, and ensuring that the model’s responses adhere to medical ethics and professional standards.

The design of the DPO dataset includes three main components: prompt (the user’s query), chosen (the desired positive sample output), and rejected (the undesired negative sample output). Specifically, the prompt represents the user’s query; the chosen represents the ideal answer the model should produce; and the rejected represents the undesirable or non-compliant answer. An example is as follows: { “prompt”: “Hello”, “chosen”: “Hello, nice to meet you”, “rejected”: “Go away, leave me alone” }. 

This section of the data was collected based on real feedback from doctors. Specifically, MedGo was tasked with responding to common questions encountered by physicians during clinical practice, and doctors provided feedback by labeling each response as either “acceptable” or “unacceptable.” Through this method, approximately 10,000 manually labeled data points were gathered to facilitate model training.

\section{Experiments}

In this study, all experiments were conducted using the PyTorch\footnote{https://pytorch.org/} framework, leveraging the HuggingFace Transformers library and the HuggingFace\footnote{https://huggingface.co/} PEFT (Parameter-Efficient Fine-Tuning) module. The experiment was conducted on 8 servers, each equipped with 8*NVIDIA A100-SXM4-80GB GPUs. This robust hardware and software setup ensured the efficient execution of training and fine-tuning processes, enabling high computational performance and scalability for our experiments. This configuration was essential for managing large-scale datasets and complex model architectures, guaranteeing reproducibility and optimal results in training MedGo.

\subsection{Results of CBLUE}
To comprehensively evaluate MedGo’s performance on various medical tasks, we conducted experiments using the CBLUE\cite{zhang2021cblue} 3.0 benchmark. CBLUE consists of 18 diverse tasks encompassing a wide range of medical text information processing requirements, including entity recognition, relation extraction, and event extraction. It also includes tasks related to medical retrieval, terminology standardization, medical text classification, semantic relation judgment of medical sentences, and advanced tasks such as medical text understanding and generation. The CBLUE benchmark provides a comprehensive evaluation framework to assess the effectiveness and generalization of MedGo in handling medical NLP tasks. The results of MedGo on the CBLUE benchmark are presented in Table \ref{tal_cblue}.

The experimental results show that MedGo has achieved compromising performance in various medical NLP tasks, including knowledge extraction, dialogue, and text classification. 
MedGo outperforms the Qwen2-72B model in tasks such as knowledge extraction, dialogue, and classification. This indicates that the model’s pre-training and fine-tuning steps effectively enhance its capabilities in the medical domain.
Since the model has not been optimized for retrieval tasks, its performance on the medical passage search task KUAKE-IR is relatively poor.
The overall score of the CBLUE evaluation benchmark is obtained by averaging the scores of all evaluation tasks—that is, by taking the macro-average of each task’s score.  Notably, the IMCS-V2-SR task provides two evaluation metrics (sentence-level F1 score and dialogue-level F1 score); both metrics are separately included in the total score calculation.

MedGo achieved the first place in the CBLUE 3.0 evaluation, even without participating in Text2DT task. This achievement shows the effectiveness of the MedGo model in Chinese medical NLP tasks.

\begin{table}[htbp]
\centering
\caption{Summary of Tasks, Evaluation Metrics, and Scores}
\begin{tabular}{|p{2.5cm}|p{7cm}|p{5cm}|p{1.5cm}|}
\hline
\textbf{Task Name} & \textbf{Task Description} & \textbf{Evaluation Metric} & \textbf{Score} \\
\hline
CMeEE & Identify named entities in Chinese medical texts and classify them into nine predefined categories. & Use strict Micro-F1 score; entities must exactly match in position and category. & 77.17 \\
\hline
CMeIE & Extract entity-relation triples from Chinese medical texts based on given relation types and schema constraints. & Use strict Micro-F1 score; predicted SPO triples must exactly match annotations. & 57.00 \\
\hline
CMedCausal & Extract causal, conditional, hierarchical, and other relations between medical concept fragments in texts. & Use Macro-F1 score; head entity, relation type, and tail entity must all be correct. & 43.88 \\
\hline
CHIP-CDEE & Extract four attributes—subject, anatomical site, descriptive term, and state—from clinical events in electronic medical records. & Use Micro-F1 score; all four attributes in the quadruple must exactly match. & 71.02 \\
\hline
CHIP-CDN & Match given original diagnostic terms to the correct standard diagnosis terms from the ICD-10 code list. & Compute F1 score based on (original diagnosis term, standard term) pairs. & 74.03 \\
\hline
CHIP-CTC & Classify short texts of Chinese clinical trial screening criteria into their semantic categories. & Use Macro-F1 score to evaluate classification performance. & 71.38 \\
\hline
KUAKE-QIC & Determine the intent category of user search queries based on medical search intent labels. & Use Accuracy to evaluate the correctness of intent classification. & 87.16 \\
\hline
CHIP-STS & Judge whether two medical-related questions have the same or similar semantics. & Use Macro-F1 score to assess semantic similarity judgment accuracy. & 86.86 \\
\hline
KUAKE-QTR & Assess the thematic matching degree between medical search queries and page titles, outputting a matching level. & Use Accuracy to evaluate matching level determination correctness. & 66.02 \\
\hline
KUAKE-QQR & Determine the semantic relationship between two medical query terms, such as exact match, subset, or unrelated. & Use Accuracy to evaluate the correctness of semantic relationship judgment. & 86.84 \\
\hline
KUAKE-IR & Retrieve relevant passages from a document pool based on the user's medical search query. & Use MRR@10 (Mean Reciprocal Rank at 10) to assess retrieval relevance. & 18.23 \\
\hline
CHIP-MDCFNPC & Extract clinical finding entities from doctor-patient dialogues and determine their status as negative, positive, or other. & Use Macro-F1 score to evaluate status determination accuracy. & 78.8 \\
\hline
IMCS-V2-NER & Recognize and classify medical entities like symptoms, tests, and procedures from doctor-patient dialogues. & Use entity-level F1 score to assess named entity recognition accuracy. & 88.72 \\
\hline
IMCS-V2-DAC & Identify the intent type of dialogue text based on doctor-patient conversations. & Use Accuracy to evaluate dialogue act classification correctness. & 83.63 \\
\hline
IMCS-V2-SR-Utterance-Level & Extract symptoms from dialogues, match to standard terms, and determine their positive or negative status. & Use sentence-level and dialogue-level F1 scores to evaluate symptom recognition and classification. & 71.83 \\
\hline
IMCS-V2-SR-Dialog-Level & Extract symptoms from dialogues, match to standard terms, and determine their positive or negative status. & Use sentence-level and dialogue-level F1 scores to evaluate symptom recognition and classification. & 74.40 \\
\hline
IMCS-V2-MRG & Generate diagnostic reports based on dialogues, including sections like chief complaint and diagnosis. & Use ROUGE metrics to assess similarity between generated and reference reports. & 57.16 \\
\hline
MedDG & Generate the doctor's next reply based on dialogue history, aiming to include correct medical entities. & Use average BLEU score and entity F1 score to evaluate reply quality and entity accuracy. & 21.68 \\
\hline
\end{tabular}

\label{tal_cblue}
\end{table}

\subsection{Results of ClinicalQA}

To verify whether MedGo can meet the actual needs of clinical doctors, we construct a high-quality dataset comprising approximately 15,000 Chinese medical consultation multiple-choice questions by medical expertors. These questions mainly originate from common issues in disease diagnosis and treatment that doctors have accumulated and organized over long-term clinical practice. Covering a wide range of medical fields, the dataset aims to reflect real medical consultation scenarios.

Each question contains four options, one of which is manually curated, while the other three are generated by invoking GPT-4. To ensure the accuracy and scientific validity of the answers, we adopted a strict double-review mechanism. Two medical students independently evaluated and initially annotated each option. If discrepancies arose in their evaluations, we involved medical experts at or above the level of associate chief physician for re-examination. These experts thoroughly assessed the disputed options, combining clinical experience and professional knowledge to decide whether to accept or replace the option. This rigorous review process ensures the high quality and credibility of the dataset.

Ultimately, we successfully constructed the ClinicalQA dataset, containing 15,000 high-quality Chinese medical consultation multiple-choice questions. This dataset is not only rich in content and highly professional but also achieves high standards in the alignment between questions and answers, providing a solid foundation for research in the field of medical natural language processing. We believe that this dataset will contribute to advancing intelligent medical applications such as medical question-answering systems and automatic diagnosis, further promoting the integration and application of artificial intelligence in the healthcare sector.

We compared the performance of three models: QWen2-72B, GPT-4o\cite{islam2024gpt}, and MedGo, with the results summarized in Table \ref{tab_qa}. The findings demonstrate that GPT-4o delivers strong performance. Furthermore, optimizing the QWen2-72B model using high-quality Chinese medical data significantly enhances its performance on this dataset. This highlights the substantial impact of leveraging specialized medical data to improve model capabilities in the medical domain.

\begin{table}[htbp]
\caption{The results of ClinicalQA.}

\centering
\begin{tabular}{|c|c|}
\hline
 Model & Accuracy \\
\hline
 QWen2-72B & 65.3 \\
 MedGo & 78.6 \\
 GPT-4o & 74.3 \\
\hline
\end{tabular}
\label{tab_qa}
\end{table}

\section{Conclusion}

We successfully developed a medical large language model, MedGo, based on extensive medical text, supervised fine-tuning data and preference alignment data. The model was built through three stages: pre-training, fine-tuning, and preference alignment. This multi-stage training strategy enables MedGo to capture complex semantics and specialized knowledge from medical texts, significantly enhancing its understanding and generative capabilities in the medical domain. To thoroughly evaluate MedGo’s practical application performance, we tested it on both the public CBLUE benchmark and our proprietary ClinicalQA dataset. The results show that MedGo achieved compromising performance across various natural language processing tasks in the medical field, including question-answering, information extraction, and clinical decision support. These outcomes confirm MedGo’s effectiveness and practicality, showcasing its potential to assist in medical practice.

In the future, we plan to incorporate additional datasets, such as MedBench\footnote{https://medbench.opencompass.org.cn/home}, to further evaluate MedGo. This will allow us to assess its performance across a broader range of medical subfields and task types. We also intend to continually expand and optimize the training data to improve the model’s accuracy and generalizability. A key future initiative is to open-source the MedGo model and the ClinicalQA dataset.

To validate the practical value of MedGo in real-world hospital settings, we have deployed it at Shanghai East Hospital\footnote{https://medgo.easthospital.cn/home}. Physicians actively use MedGo in their daily clinical routines to assist in decision-making and medical inquiries. MedGo’s generated responses are continuously evaluated by doctors, who provide feedback on their accuracy and relevance. At present, MedGo collects approximately 1,000 feedback entries per day, encompassing various clinical interactions. This ongoing feedback loop is crucial for fine-tuning MedGo’s performance, allowing for iterative improvements in its response accuracy and relevance. This deployment and feedback system ensure that MedGo consistently meets the rigorous standards required in clinical environments, thereby enhancing its practical application in healthcare settings.

\bibliographystyle{unsrt}  
\bibliography{template}  %%% Remove comment to use the external .bib file (using bibtex).


% George Kour and Raid Saabne.
% \newblock Real-time segmentation of on-line handwritten arabic script.
% \newblock In {\em Frontiers in Handwriting Recognition (ICFHR), 2014 14th
%   International Conference on}, pages 417--422. IEEE, 2014.

% George Kour and Raid Saabne.
% \newblock Fast classification of handwritten on-line arabic characters.
% \newblock In {\em Soft Computing and Pattern Recognition (SoCPaR), 2014 6th
%   International Conference of}, pages 312--318. IEEE, 2014.

% \bibitem{hadash2018estimate}
% Guy Hadash, Einat Kermany, Boaz Carmeli, Ofer Lavi, George Kour, and Alon
% \newblock Estimate and replace: A novel approach to integrating deep neural
%   networks with existing applications.
% \newblock {\em arXiv preprint arXiv:1804.09028}, 2018.

% \end{thebibliography}

\end{document}